\documentclass{article}
\usepackage{graphicx}
\usepackage{subcaption}

\usepackage{arxiv}

\usepackage[utf8]{inputenc} 
\usepackage[T1]{fontenc}    
\usepackage{hyperref}       
\usepackage{url}            
\usepackage{booktabs}       
\usepackage{amsfonts}       
\usepackage{nicefrac}       
\usepackage{microtype}      
\usepackage{lipsum}		
\usepackage{graphicx}
\usepackage{natbib}
\usepackage{doi}

\title{A Multi-Modal Approach for Face Anti-Spoofing in Non-Calibrated Systems using Disparity Maps}

\author{ \href{https://orcid.org/0000-0002-5006-9300}{\includegraphics[scale=0.06]{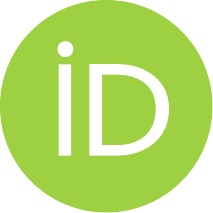}\hspace{1mm}Ariel Larey} \\
	RealSense\\
	\texttt{ariel.lahrey@intel.com} \\
	\And
	\href{https://orcid.org/0009-0005-8619-0443}{\includegraphics[scale=0.06]{orcid.pdf}\hspace{1mm}Eyal Rond} \\
	RealSense\\
	\texttt{\tt\small eyal.rond@intel.com} \\
        \And
        \href{https://orcid.org/0009-0002-1669-4679}{\includegraphics[scale=0.06]{orcid.pdf}\hspace{1mm}Omer Achrack}\thanks{Corresponding author\\ Authors contributed equally to this work} \\
	RealSense\\
	\texttt{omer.achrack@intel.com} \\
}

\renewcommand{\headeright}{Larey et al.}
\renewcommand{\undertitle}{}
\renewcommand{\shorttitle}{Face Anti-Spoofing in
Non-Calibrated Systems}

\hypersetup{
pdftitle={A Multi-Modal Approach for Face Anti-Spoofing in Non-Calibrated Systems using Disparity Maps},
pdfsubject={q-bio.NC, q-bio.QM},
pdfauthor={Ariel Larey, Eyal Rond, Omer Achrack},
pdfkeywords={Face recognition, Anti-spoofing, non-calibrated systems, disparity maps},
}

\begin{document}
\maketitle

\begin{abstract}
Face recognition technologies are increasingly used in various applications, yet they are vulnerable to face spoofing attacks. These spoofing attacks often involve unique 3D structures, such as printed papers or mobile device screens. Although stereo-depth cameras can detect such attacks effectively, their high-cost limits their widespread adoption. Conversely, two-sensor systems without extrinsic calibration offer a cost-effective alternative but are unable to calculate depth using stereo techniques.
In this work, we propose a method to overcome this challenge by leveraging facial attributes to derive disparity information and estimate relative depth for anti-spoofing purposes, using non-calibrated systems. We introduce a multi-modal anti-spoofing model, coined Disparity Model, that incorporates created disparity maps as a third modality alongside the two original sensor modalities. We demonstrate the effectiveness of the Disparity Model in countering various spoof attacks using a comprehensive dataset collected from the Intel® RealSense™ ID Solution F455. Our method outperformed existing methods in the literature, achieving an Equal Error Rate (EER) of 1.71\% and a False Negative Rate (FNR) of 2.77\% at a False Positive Rate (FPR) of 1\%. These errors are lower by 2.45\% and 7.94\% than the errors of the best comparison method, respectively. Additionally, we introduce a model ensemble that addresses 3D spoof attacks as well, achieving an EER of 2.04\% and an FNR of 3.83\% at an FPR of 1\%. Overall, our work provides a state-of-the-art solution for the challenging task of anti-spoofing in non-calibrated systems that lack depth information.
\end{abstract}


\section{Introduction}
\label{sec:intro}

In recent years, face recognition technologies have seen a significant increase in popularity across a wide line of products, including access control systems, phone unlocking mechanisms, digital payment platforms, and attendance tracking systems.
Despite their widespread adoption, these systems remain susceptible to substantial security risks, such as Face-Spoofing (FS) or Face Presentation Attacks (FPA). Consequently, detecting spoof attempts in modern face recognition systems has become an active area of research in both academia and industry. As a result, numerous papers and datasets have been introduced in this field (\cite{CasiaSurf, steiner2016reliable}), and anti-spoofing modules are increasingly being integrated into modern products prior to the face recognition phase (See Appendix \ref{app:fa_pipeline} for more details).

The domain of spoof attacks can be categorized into two primary types: 2D attacks (e.g., printed papers, different types of screens) and 3D attacks (e.g., 3D masks made from rigid materials, silicone masks, fabric masks). For 2D attacks, stereo-depth cameras have demonstrated promising results (\cite{zhang2019dataset}). However, stereo-depth cameras could be expensive and less feasible for widespread adoption in most real-life scenarios.
On the other hand, Near-IR (NIR) sensors are less-expensive than stereo-depth cameras, and are used as a preferable option for lower-budget use cases and for large-scale distributed edge computing systems. Additionally, NIR sensors have also demonstrated high performance in detecting 3D FAS (e.g., masks) as reported in \cite{jiang2019multilevel}.
Another approach to address the FAS problem is the development of multi-modality systems. In this type of system, decisions are based on several modalities (e.g., visual information, NIR information, depth information or even temporal information from videos), which model different aspects of the physical environment. Utilizing multiple modalities can enhance robustness against various types of spoof attacks in real-world scenarios (\cite{lin2024suppress}).

Our research is based on a relatively new and promising product in the market: Intel® RealSense™ ID Solution F455. This product features two NIR sensors coincides with the multi-modality approach, since the sensors utilize different color schemes, thereby enriching the texture information compared to a single color scheme system \footnote{For further information about the exact sensor technical specifications, please visit the official website of the product: \url{https://store.intelrealsense.com/intel-realsense-id-solution-f455-peripheral-version-with-software.html}}.

Moreover, the systems’ sensors lack extrinsic calibration, which makes the product more affordable and suitable for lower-cost and scalable security systems. However, the absence of rectification information prevents the calculation of depth using stereo techniques. In this work, we introduce a unique method to overcome the lack of camera parameters knowledge, by leveraging facial attributes to derive disparity information and estimate relative depth for the anti-spoofing task. We train an anti-spoofing model using a multi-modal approach, where the model incorporates disparity information as an effective third modality, in addition to the two physical NIR sensor modalities.

An additional critical consideration for face recognition systems is data privacy. Various solutions in cloud computing address this concern, but our model is designed to operate in an edge-compatible manner. This approach effectively prevents the leakage of sensitive data, as the entire face recognition process is conducted on the edge device itself. Furthermore, this method offers the advantage of functioning independently of network connectivity, thereby making it suitable for a broader range of use cases. To meet the requirements for edge computing, our model is small in terms of memory footprint and with the ability to operate with lower precision.
\newline To summarise, our contributions are as follows:
 
\begin{itemize}
   \item Develop a novel method for extracting disparity information as a new virtual depth sensor data.
   \item Design and implement an anti-spoofing model that integrates three different modalities to achieve superior performance in detecting spoofing attacks, particularly with non-calibrated sensor inputs.
   \item Create multi modality solution for FAS system suitable for edge devices.
\end{itemize}

\begin{figure}
    \begin{center}
    \begin{subfigure}[t]{0.225\columnwidth}  \includegraphics[width=0.9\linewidth]{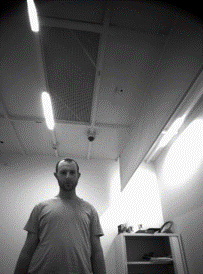}
        
    \end{subfigure}
     \begin{subfigure}[t]{0.225\columnwidth}
        \includegraphics[width=0.9\linewidth]{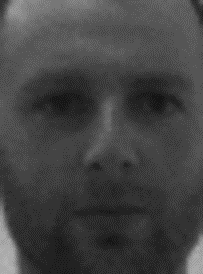}
        
    \end{subfigure}
    \begin{subfigure}[t]{0.225\columnwidth}
        \includegraphics[width=0.9\linewidth]{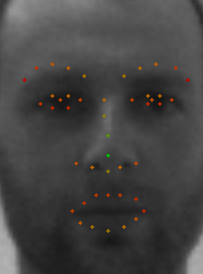}
        
    \end{subfigure}
    \begin{subfigure}[t]{0.225\columnwidth}
        \includegraphics[width=0.9\linewidth]{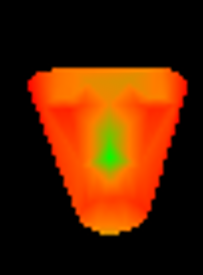}
        
    \end{subfigure}
    \begin{subfigure}[t]{0.225\columnwidth}  \includegraphics[width=0.9\linewidth]{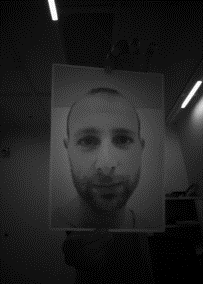}
        \caption{}
    \end{subfigure}
     \begin{subfigure}[t]{0.225\columnwidth}
        \includegraphics[width=0.9\linewidth]{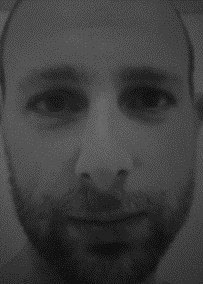}
        \caption{}
    \end{subfigure}
    \begin{subfigure}[t]{0.225\columnwidth}
        \includegraphics[width=0.9\linewidth]{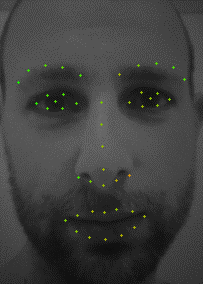}
        \caption{}
    \end{subfigure}
    \begin{subfigure}[t]{0.225\columnwidth}
        \includegraphics[width=0.9\linewidth]{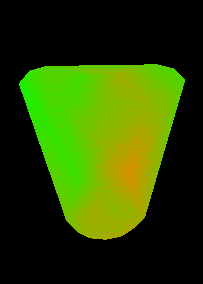}
        \caption{}
    \end{subfigure}
        \caption{Demonstration of disparity maps creation pipeline for live (top) and spoof (bottom) examples. First, image acquisition is performed by the device (a). Next, Face Detection is applied (b) followed by Facial landmarks extraction and sparse disparity calculation (c). Finally, disparity maps are produced by linear interpolation (d). Green represents higher disparity than the red.}
        \label{fig:real_spoof_disp}
    \end{center}
\end{figure}

\section{Related work}
FAS is an active research area in both industry and academia, where various sensors are examined and utilized to counter different spoofing attacks.

\paragraph{Multi-Modality FAS} Multi-modality is a robust mechanism that enables FAS systems to integrate various physical attributes from different sensors. \cite{jiang2019multilevel} utilized visible light and NIR sensors with three levels of data fusion. In the Data Level Fusion, data from both modalities are concatenated across the channels and are used as input to a Convolutional Neural Network (CNN). In the Feature Maps Level Fusion, feature maps are computed separately for each modality via a CNN, then they are concatenated across the channels and are used as input to a sub-network classification module. In the Fully Connected Level Fusion, fusion is applied only to the final layers of the network. The final prediction probability is a weighted average of the probabilities from all three fusion sub-models. 
In \cite{CasiaSurf}, the authors utilized active stereo depth input, comprising RGB and NIR data. They integrated these inputs into a single decision at the CNN level using the Squeeze and Excitation (SE) mechanism (\cite{hu2018squeeze}).
In \cite{lin2024suppress}, the authors used Visual Transformer (ViT) blocks as a backbone for each modality (RGB and NIR). They also introduced a decision mechanism based on uncertainty estimation, with the authors adaptation of Monte Carlo sampling (\cite{kendall2015bayesian}) to evaluate the feature unreliability in each modality. Based on the uncertainty score, they weighted the prediction in cross-modal attention fusion.
\cite{yu2020multi} proposed using Central Difference Convolution (CDC) (\cite{yu2020searching}) to enhance the model's ability to capture intrinsic live or spoof features from each modality separately. Subsequently, the features are concatenated and processed by a single sub-model that produces a binary mask, where the final score is determined by averaging the values of the mask.

\begin{figure*}
  \centering
  \includegraphics[width=1\textwidth]{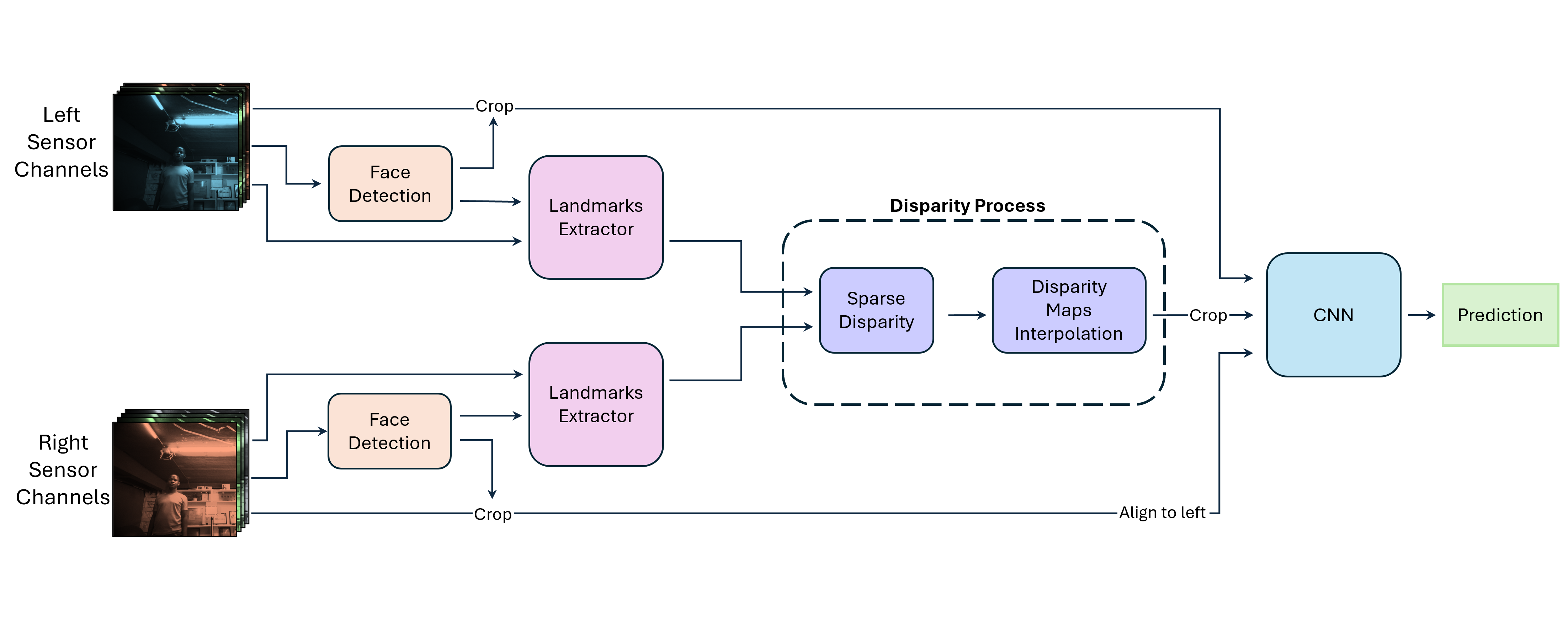}
  \caption{Disparity Model full pipeline. Initially, faces are detected, and facial landmarks are extracted from the data of both sensors. These landmarks serve as key points for sparse disparity calculation along both the horizontal and vertical axes. This is followed by spatial interpolation, resulting in two disparity maps. Finally, the aligned right sensor crop, the left crop, and the disparity maps are concatenated and processed by a CNN to predict whether the input is a spoof or live.}
  \label{fig:disp_pipeline}
\end{figure*}

\paragraph{Depth-based FAS} The primary motivation for employing depth sensing in FAS systems is its robustness against 2D spoof attacks, which are the most common attacks in real-life scenarios. \cite{atoum2017face} utilized two streams during the process. The first stream employs a trained depth estimator using images extracted from a single source, while the second stream processes patches cropped from the RGB input. Finally, a unified decision is determined via a classification sub-model that employs a weighting mechanism between the scores from both streams. 
In \cite{wang2017robust}, depth images captured using a Kinect device were utilized. The authors employed a CNN architecture, followed by a linear Support Vector Machine (SVM), to classify the extracted features into live or spoof categories. 
In \cite{zhang2022advanced}, a PSPNet architecture (\cite{zhao2017pyramid}) was trained to generate segmented depth maps. A decision rule was then applied to these predicted depth maps to determine whether the input was live or a spoof attack.

\paragraph{Temporal-based FAS} Video-based FAS systems can capture liveness by analyzing motion patterns across the video, a capability that single-image FAS systems lack. \cite{yang2019face} designed a Spatio-Temporal Anti-Spoofing Network (STASN), which comprises distinct sub-networks for spatial and temporal components. The authors augmented the spoof data by applying various noise patterns and transformations to their collected live samples, making them resemble spoof samples. \cite{li2016generalized} generated temporal feature vectors from cropped face regions of interest (ROIs) across the video. These vectors were calculated using temporal filtering and Fast Fourier Transform (FFT) and subsequently were classified by a Support Vector Machine (SVM). \cite{ge2020face} employed a CNN combined with Long Short-Term Memory (LSTM, \cite{greff2016lstm}) for anti-spoofing predictions from videos. They utilized Eulerian motion magnification (\cite{wu2012eulerian}) as part of the network’s input process, to enhance subtle motions of live faces.
\cite{anthony2022active} developed a dataset that enables the extraction of live or spoof information from motion. They captured videos for each sample at two known distances. Similar to our approach, they calculated distances between facial features extracted from two images. A feature vector for each sample was then created from these computed distances, and was classified using various deep learning and classical machine learning methods. In our work, we demonstrate that this approach is insufficient for the challenging task of anti-spoofing using non-calibrated sensors. To address this domain challenges, we introduce a method that leverages the raw distances of the facial features, into spatial proxy-depth information as an additional input to the anti-spoofing model.

\section{Method}

\subsection{Camera sensors process}
Standard RGB cameras are designed to produce visually appealing images based on human visual perception. Thus, various Digital Signal Processors (DSPs) are employed to process raw images and enhance their visual appeal. However, in Facial Anti-Spoofing (FAS) systems, the objective is fundamentally different. The primary goal is to extract features pertinent to distinguish between live and spoof images, rather than generating visually appealing images. Therefore, in our pipeline, the demosaicing algorithm (\cite{li2008image}) is not performed, and instead the raw Bayer pattern from the sensor, reshaped into four channels, is used. Specifically, two four-channel images are produced: one from the left sensor ($s_l$) and one from the right sensor ($s_r$). Further details are provided in Appendix \ref{app:bayer_vs_bggr}.

\subsection{Disparity maps}
Certain forms of Spoof attacks are characterized by a unique facial structure. For instance, two-dimensional spoof attacks are composed of a plane within a sphere, whereas alternative attacks might be performed via other geometric structures, such as a cylinder. Occasionally, the information provided by texture alone is insufficient for the precise operation of anti-spoofing systems, necessitating the incorporation of three-dimensional geometric knowledge. In response to the absence of depth data in the system’s ASIC, and the lack of intrinsic and extrinsic information from the sensors, we acquire three-dimensional knowledge through a specialized methodology.

Two images are under consideration, the initial one being derived from the left sensor ($s_l$), and the subsequent one from the right sensor ($s_r$). Both images are applied to a Face Detector (FD) model followed by a Facial landmarks extractor (LE). The latter yields 45 facial landmarks distributed across the face. Both the FD and LE models are specifically trained to process data from the Intel® RealSense™ ID Solution F455 and are dedicated to this domain. For additional details, please refer to Appendix \ref{app:Supplementary_Models}.
Formally, this can be expressed as:
\begin{equation}
rect_k = FD(s_k)
\end{equation}

\begin{equation}
lms_k = LE(rect_k, s_k)
\end{equation}

Where $k\in\{l,r\}$ refers to the sensor, $rect_k$ is the rectangle circumscribing the face captured by sensor $k$, and $lms_k$ represents 45 facial landmarks extracted from sensor $k$ image, while each facial landmark is a two-dimensional point within the image.

To estimate relative depth, we utilize pairs of landmarks from two sensors, each projected from the same 3D point on the sphere. For a given point on the face, denoted as $P_i$, and its corresponding facial landmarks from each sensor, $lms_{l,i}$ and $lms_{r,i}$ we can determine the disparity in each dimension by subtracting the former from the latter:

\begin{equation}
disp_{i,x}=lms_{r,i,x}-lms_{l,i,x}
\end{equation}

\begin{equation}
disp_{i,y}=lms_{r,i,y}-lms_{l,i,y}
\end{equation}

Where $i\in\{1,45\}$ is the facial landmark index, $x$ and $y$ represent the horizontal and vertical dimension within the image respectively, and $disp_{i,d}$ is the disparity of facial landmark $i$ over dimension $d$. 

Understanding disparity, the difference in image location of an object seen along two different lines of sight, can provide valuable insights into the depth relationships between facial features. For instance, in a capture of live instance in frontal pose, the tip of the nose should exhibit greater disparity than the pupil of the eye, given that the nose is closer to the camera system, unlike in some 2D spoof attacks.
Another significant advantage of understanding disparity is the capability to estimate the depth of the entire head in relation to its size. This becomes particularly crucial when attempting to detect spoof attacks that are often carried out using smaller images.
In calibrated stereo systems, disparity is typically computed using predefined system information. However, in our approach, we have calculated it utilizing the knowledge of facial attributes.

Finally, to complete the spatial disparity information of the face, we perform a linear interpolation (LI) across the pixels of the left sensor. This process is guided by the sparse disparity values derived from the facial landmarks.

\begin{equation}
dispmap_x = LI(lms_{l,x}, disp_{x})
\end{equation}

\begin{equation}
dispmap_y = LI(lms_{l,y}, disp_{y})
\end{equation}

Where $x$ and $y$ represent the horizontal and vertical dimensions of the image respectively. $dispmap_{d}$ is the disparity-map calculated using the sparse disparity ($disp_{d}$) over dimension $d$. The linear interpolation is implemented using the SciPy \footnote{For further information about the exact Scipy interpolation implementation: \url{https://docs.scipy.org/doc/scipy/reference/generated/scipy.interpolate.griddata.html}} library (\cite{2020SciPy-NMeth}). \autoref{fig:real_spoof_disp} presents live and spoof examples of disparity maps and their process steps.

\subsection{Disparity Model}
\paragraph{Data-process}
A dedicated deep learning model to predict 2D spoof attacks is developed by utilizing relative disparity knowledge. Given that the disparity map contains spatial pixel-wise data, it can be processed by convolutional neural network (CNN) kernels along with their corresponding texture maps on a pixel-wise basis. Specifically, the model’s input comprises ten channels from three modalities: four channels from the left sensor, four channels from the right sensor, and two channels are the disparity maps both for horizontal and vertical dimensions.
The complete multi-modal data process is illustrated in \autoref{fig:disp_pipeline}. The four-channel images from both sensors are first applied to a face-detection model and cropped around the detected face with extension of 15\% per dimension. The image from the right sensor is then aligned to the left crop with four degrees of freedom in terms of scale and translation. Subsequently, the disparity maps, created based on the spatial structure of the left sensor, are cropped and concatenated with the two sensors crops. Finally, the 10-channels input is resized to 128x128 pixels and is fed into the disparity model.

\begin{figure}
  \centering
  \includegraphics[width=0.8\columnwidth]{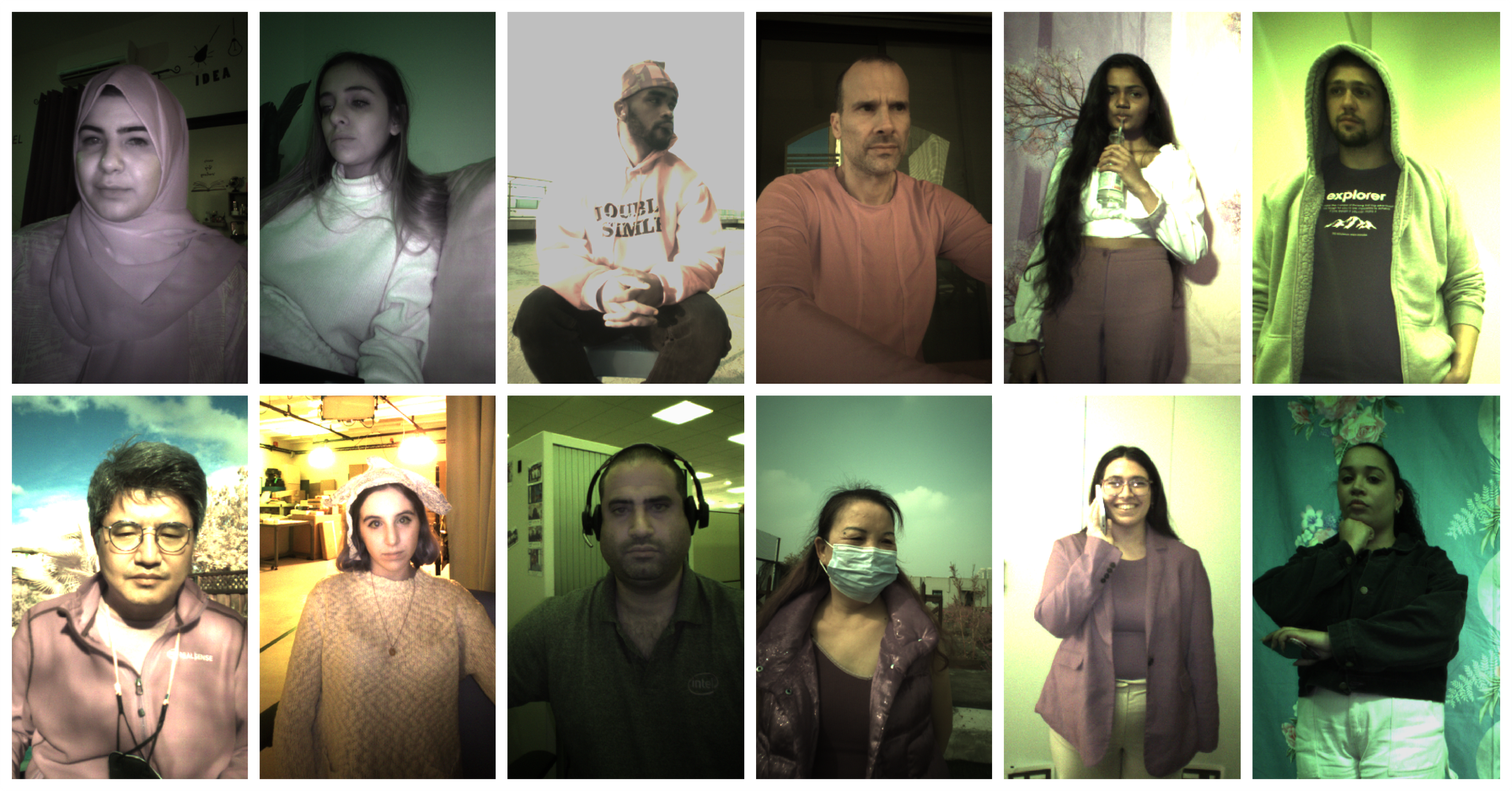}
  \caption{Examples for live samples in the collected data.}
  \label{fig:live_examples}
\end{figure}

\begin{figure}
  \centering
  \includegraphics[width=0.8\columnwidth]{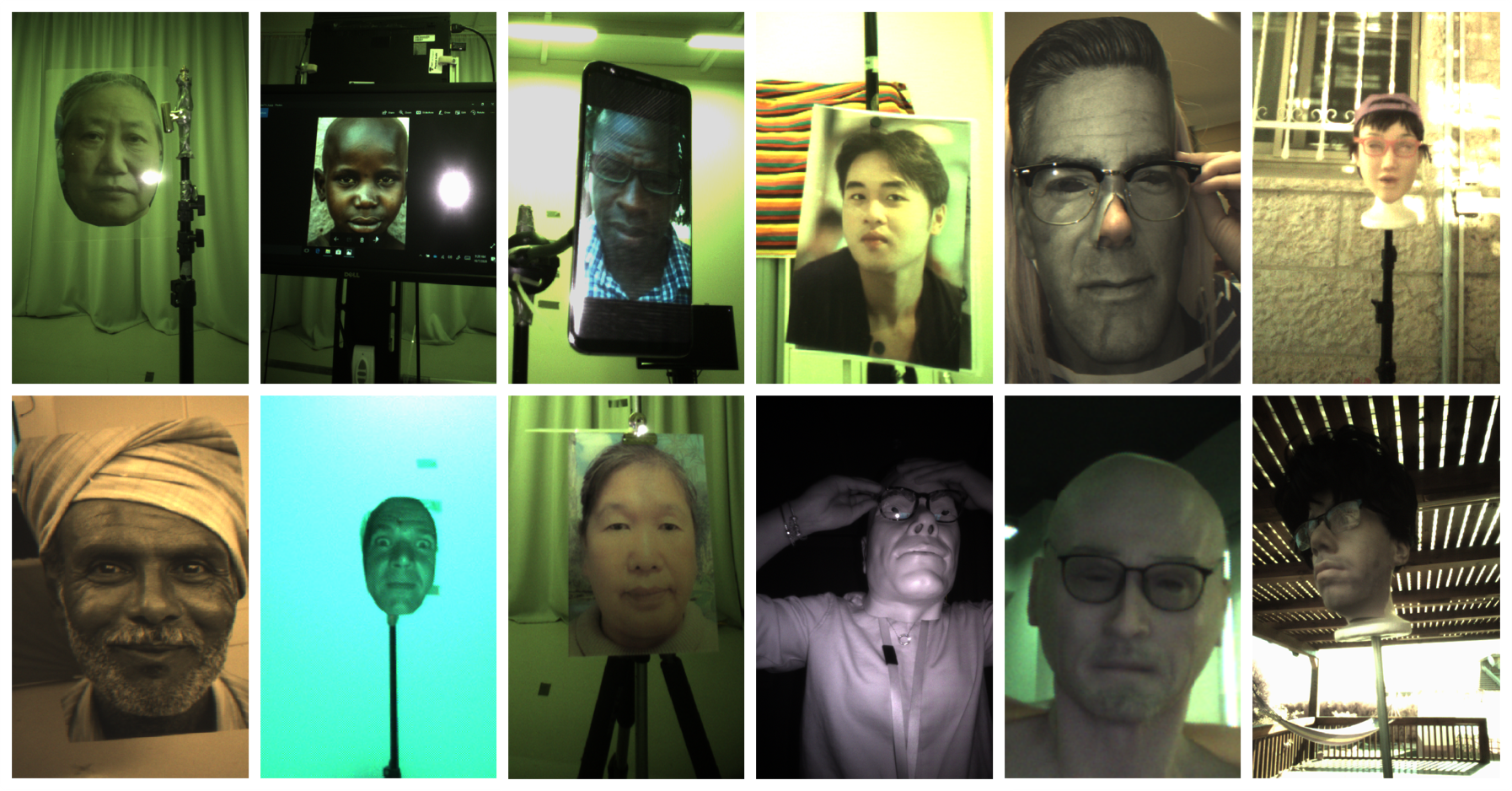}
  \caption{Examples for spoof samples in the collected data.}
  \label{fig:spoof_examples}
\end{figure}

\paragraph{Model Details}
To achieve optimal product efficiency and the ability to operate on an edge device, the disparity model is based on a modified version of MobileNetV2 architecture (\cite{sandler2018mobilenetv2}). The Disparity model comprises 27 convolution layers, each followed by a batch normalization layer and a ReLU activation function. This model contains approximately 1 million trainable parameters.

In security systems it is essential to obtain reliable predictive probabilities. This is particularly crucial when integrating multiple models, each with distinct probability decision thresholds. However, the conventional SoftMax loss function tends to exhibit overconfidence, which subsequently results in outputs that are unreliable as probability distributions (\cite{sensoy2018evidential, achrack2020multi, neumann2018relaxed}). To address this issue, a high Evidential Deep Learning loss is employed during the training of the model, thereby ensuring the generation of stable and well-calibrated outputs (\cite{sensoy2018evidential}).

\paragraph{Data Augmentations}

The disparity model employs a multi-modal approach, which results in a unique methodology for data augmentations during training. This technique comprises three levels of augmentation:

\begin{itemize}

\item Landmarks Augmentations. Augmentations are performed on facial landmarks to enhance stability against landmark extractor errors, which could lead to errors in the disparity maps. These augmentations include noising random number of landmarks, with random translation values per landmark.
Another type of facial landmark augmentation simulates landmark extractor bias for an entire facial organ region (e.g., a shift in nose landmarks coordinates). For this augmentation, a facial organ is uniformly sampled, and all landmarks in that region are translated by the same magnitude. In both augmentations, the magnitude of translation is determined by a uniform sampling method, with a maximum limit set to 6\% of the respective facial dimension.
Additionally, we introduce the outlier augmentation, where the magnitude of the translation is uniformly sampled between 6\% to 14\% of the face size per dimension. In this scenario, up to four facial landmarks are translated with a larger magnitude to simulate outliers.

\item Sensor Intensity Augmentations. To enhance robustness against a range of variables that influence the intensity of sensor outputs, a series of standard augmentations are applied directly to the output channels of these sensors. These augmentations include color jittering and mean-variance adjustments, which modify the brightness and contrast within the image. Additionally, noise is introduced to alter the intensity of individual pixels across the image, and motion blur is applied to mimic the blurring effect caused by head movements. It is important to note that these augmentations are exclusively applied to the direct outputs of the sensors, and not to the disparity maps.

\item Spatial Augmentations. These augmentations, which include horizontal flipping, rotation, and shearing, are designed to effectively expand the training dataset. Furthermore, a cut-out augmentation technique is employed, wherein a specific region of the input is masked to minimize the model’s dependency on particular regions during training. The ratio between the face and the background size is also augmented, and is being uniformly sampled between 40\% to 80\% during the cropping process. Additionally, the bounding box of the face undergoes a random translation, up to 20\% of its size, prior to the cropping process. This is done to enhance the model’s robustness against the face detector ‘s errors. The same set of spatial augmentations is applied across all modalities including the disparity maps.

\end{itemize}

\subsection{Model Ensemble setup}
The disparity model is designed to counteract unique structures of spoof attacks, particularly 2D attacks, by using additional proxy-depth knowledge. However, the disparity maps can mislead the model in 3D attacks scenarios where face geometry closely mimics reality. To address this, two additional models are trained to address 3D attacks.

The first additional model, referred to as the Left Model, processes inputs from the left sensor’s four channels. The second model, coined as the Right Model, is designed to process data from the right sensor’s four channels. Both these models share the same architectural design as the Disparity Model, with the exception of the input layer size. They also employ the same data augmentation techniques, barring the landmarks augmentations.
Eventually, these three models are integrated into a 3-model-ensemble. The final decision is determined by an ‘OR’ operation applied to the spoof predictions. I.e. if any one of the models predicts a spoof attack for a given system input, the ensemble prediction is classified as a spoof attack.
 
Each model’s decision threshold is determined using a dedicated software that seeks an optimal threshold combination among the models. This search is conducted in two stages: initially at high level with high granularity iterations, followed by a more detailed search with smaller granularities, centered around the results from the high-level search. This approach ensures a comprehensive and precise threshold determination.

\begin{figure}
  \centering
  \includegraphics[width=0.9\columnwidth]{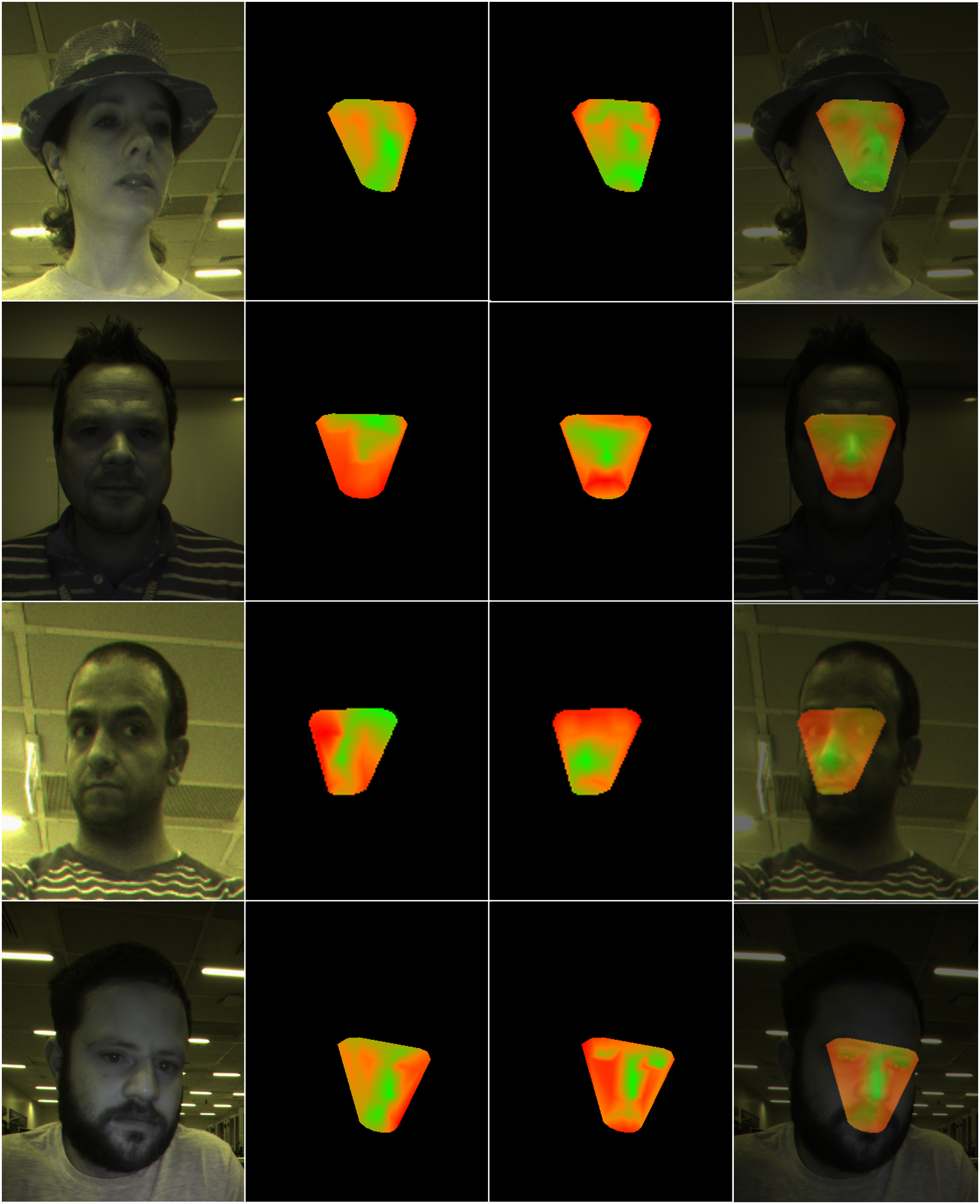}
  \caption{Examples for Disparity maps produced from live samples.}
  \label{fig:live_examples_disp}
\end{figure}

\section{Experiments}
\subsection{Data}
In this study, we demonstrate our multi-modal approach, using data collected from the Intel® RealSense™ ID Solution F455 NIR sensors, which comprises two non-calibrated sensors. The data was captured using 20 distinct systems, to avoid over-fitting to a single device. 
The collected live samples contained 10.7K different identities in various conditions, with total amount of approximately 1.7M images. These conditions included diverse background lighting, distances from the device (up to 2 meters), head poses (up to 40 degrees for yaw, pitch, and roll), locations (both indoors and outdoors), occluding accessories (such as hats, sunglasses, and face COVID-19 masks), and facial expressions (\autoref{fig:live_examples}).

Moreover, variety of spoof attacks were collected. The 2D attacks involved printed papers of different sizes (A3, A4), television screens, and mobile phone screens as well as curved screens. In each 2D spoof attack, a face that fulfilled the real footage criteria was displayed. The 3D attacks incorporated cut papers placed over a live instance’s face that revealed its nose, papers rolled into a cylinder, and latex masks worn by live individuals. In total, 2M various spoof data were collected (\autoref{fig:spoof_examples}). 

The dataset was divided such that 85\% was allocated for training and 15\% for validation. Additionally, an external test set comprising approximately 100,000 samples of both live and spoof attacks was collected separately to enhance the generalization of the evaluation. It is important to note that each dataset contained different individual subjects. In the test and validation datasets, the proportion of 2D attacks within the total spoof data is larger than that of 3D attacks. This is based on the belief that 2D attacks more accurately reflect real-life scenarios, as they are easier to produce using simple means.

\begin{figure}
  \centering
  \includegraphics[width=0.9\columnwidth]{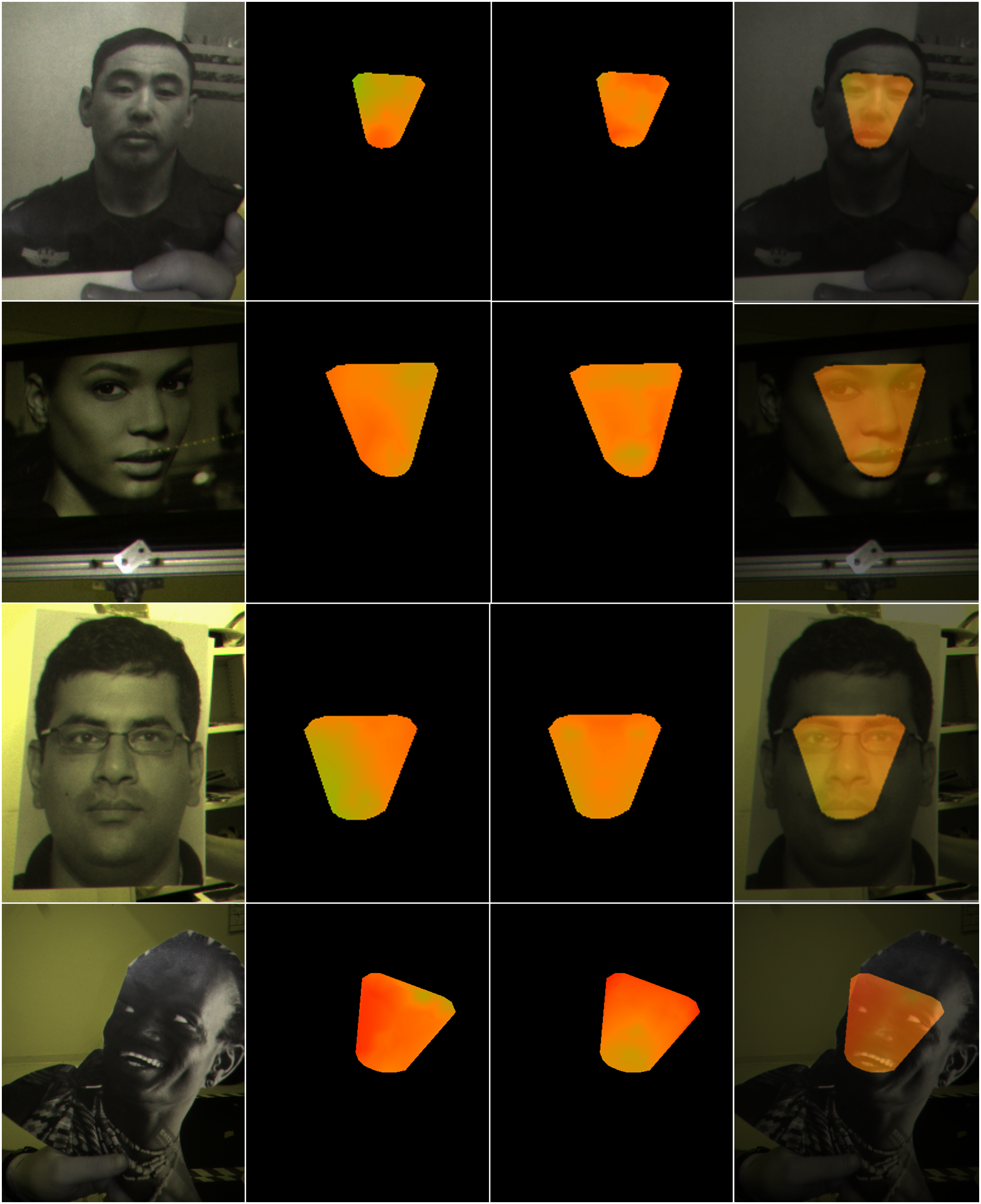}
  \caption{Examples for Disparity maps produced from 2D spoof attacks samples.}
  \label{fig:spoof_examples_disp}
\end{figure}

\subsection{Metrics}
In our convention, live samples are classified as positive instances, while samples of spoof attacks are classified as negative instances. Consequently, the False Negative Rate (FNR) represents the proportion of failures among all live instances. On the other hand, the False Positive Rate (FPR) signifies the proportion of False predictions among all spoof attack instances. 

\begin{table}[ht!]
\centering
 \caption{Different methods comparison for live and 2D attacks}
  \label{tbl:main_results}
{%
\begin{tabular}{lccccc}
\toprule
& EER     & \shortstack{FNR\\FPR=0.05} & \shortstack{FNR\\FPR=0.03} & \shortstack{FNR\\FPR=0.01} \\
 \cmidrule(r){1-5}
Facial Features Pairs & 34.45\% & 73.17\% & 78.93\% & 85.69\% \\
CNN & 6.47\%  & 7.79\% & 10.83\% & 18.89\%  \\
Patch-CNN & 5.87\% & 6.84\% & 9.89\% & 16.31\% \\
Multilevel fusion & 4.16\% & 3.99\% & 4.39\% & 10.71\% \\
Disparity Model & \textbf{1.71\%} & \textbf{0.64\%} & \textbf{1.12\%} & \textbf{2.77\%} \\
 \bottomrule
\end{tabular}
}
\\
\scriptsize Lower values indicate better performance.
\end{table}

The primary metric for assessing the models is the Equal Error Rate (EER), which is the error rate at which both FNR and FPR are equal. This approach ensures a balanced evaluation of the model’s performance in identifying both positive and negative instances.

For product-oriented evaluation of the models, additional metrics are employed. In these cases, the decision thresholds are calibrated to ensure the model predicts a predefined FPR on the test benchmark. Once the FPR is fixed, the model’s performance is assessed based on its FNR at the corresponding threshold. This evaluation is conducted for FPR values of 5\% ($\frac{5}{100}$ miss acceptance rate), 3\% ($\frac{3}{100}$ miss acceptance rate), and 1\% ($\frac{1}{100}$ miss acceptance rate).

\subsection{Experimental Setup}
The disparity model is specifically designed to counter 2D attacks by leveraging additional proxy-depth knowledge. Consequently, we train and evaluate it using only live and 2D attacks samples. As baselines, we implemented four other anti-spoofing approaches inspired by the literature.

The first approach, termed Facial Features Pairs, involves calculating the Euclidean distance between all pairs of facial landmarks (\cite{anthony2022active}), resulting in a 2025-feature vector derived from 45 facial landmarks of each sensor source. Several multi-layer perceptrons (MLPs) were examined using the feature vectors as inputs, and we report the results for the MLP with the best performance.

Additionally, we trained a CNN model that processes data from both sensors as an 8-channel input. This model shares the same architecture as the disparity model, except for the first input layer. Another approach examined is the Patch-CNN, where patches of the concatenated data are processed by a shared CNN, and the final prediction is averaged among all patch outputs, as described in \cite{atoum2017face}.

Furthermore, we implemented the Multilevel Fusion model architecture from \cite{jiang2019multilevel}, adjusting the model size to approximately 1 million trainable parameters, similar to the size of the disparity model. In the original paper, the authors integrated data from Visual and NIR sensors at three steps of the deep learning pipeline (\cite{jiang2019multilevel}). Here, we perform the same integration, but by using data from the two NIR sensors of the Intel® RealSense™ ID Solution F455 device.

All models were trained on the same dataset using the TensorFlow framework (\cite{abadi2016tensorflow}) and two GeForce GTX 1080 Ti GPUs. We employed the ADAM optimizer (\cite{kingma2014adam}) with a momentum of 0.9 and a learning rate of 1e-4. The training process involved a batch size of 32 over 20 epochs, where the checkpoint with the highest accuracy on the validation set was reported.

Finally, we trained the Left and Right models as complementary components of the model ensemble, specifically designed to address 3D attacks. These models share the same architecture and training procedure as the Disparity Model, with the addition of incorporating instances of 3D attacks into the dataset.

\section{Results}

\begin{figure}
  \centering
  \includegraphics[width=0.5\columnwidth]{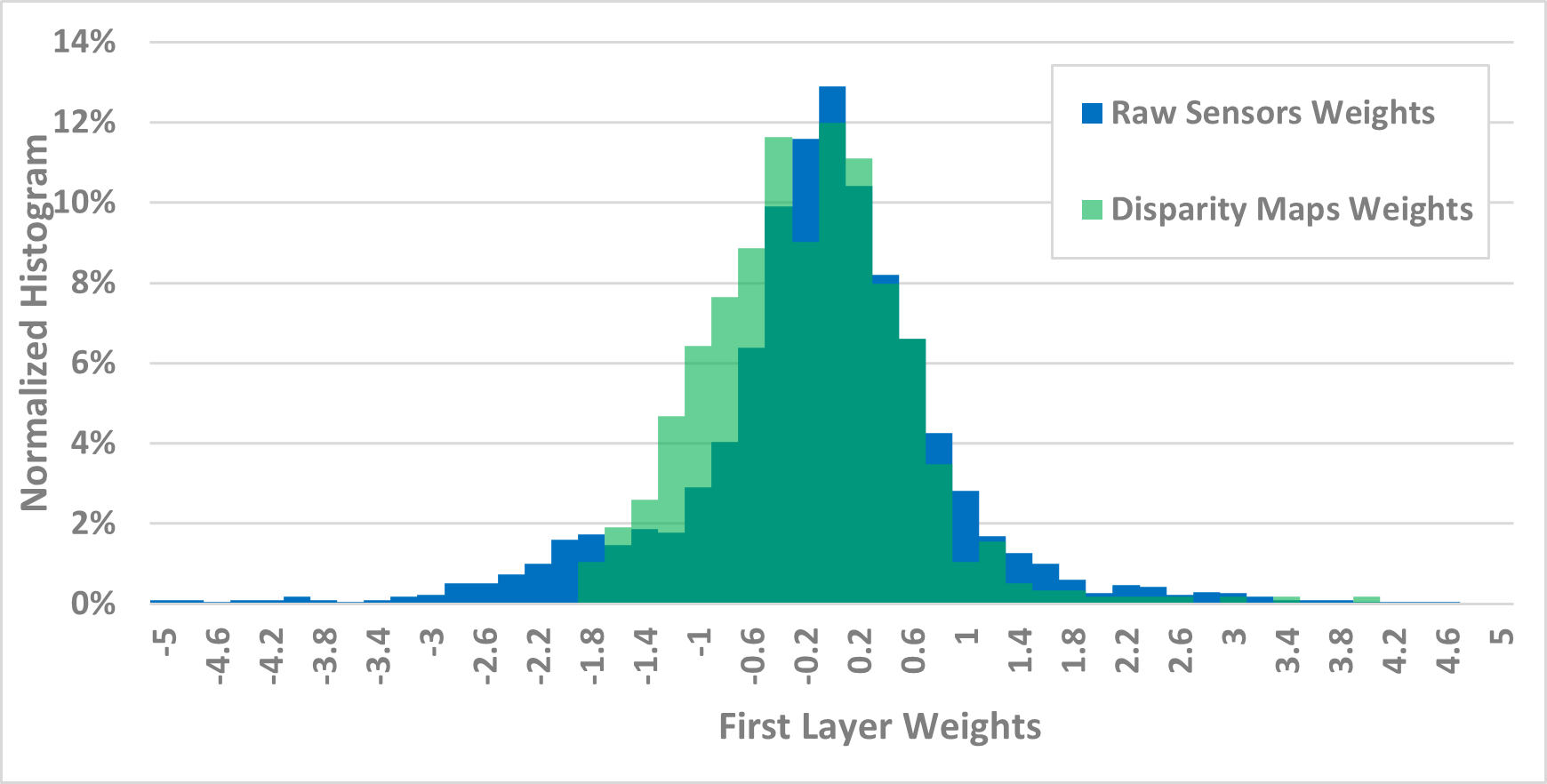}
  \caption{First Layer's weights distribution}
  \label{fig:weights}
\end{figure}

\subsection{Disparity Maps Examples}
\autoref{fig:live_examples_disp} and \autoref{fig:spoof_examples_disp} illustrate additional examples of live and spoof disparity maps, respectively. In these figures, each row corresponds to a different subject. The leftmost column displays the original image. Moving from left to right, the subsequent columns show the disparity maps generated along the horizontal axis ($dispmap_x$) and the vertical axis ($dispmap_y$). For ease of visualization, each disparity map is color-coded, with green indicating higher disparity values than red. Finally, to highlight the three-dimensional structure of the disparity spatially, both disparity maps are overlayed on the original image in the rightmost column. As observed, the live disparity maps replicate the three-dimensional structure of the subject, whereas the disparity in two-dimensional spoof attacks exhibits a planar structure with linear spatial variations.

\begin{table}[ht!]
\centering
 \caption{Disparity Model Ablation Study}
  \label{tbl:ablation}
{%
\begin{tabular}{ccccccc}
\toprule
\multicolumn{3}{c}{Components} & &\multicolumn{3}{c}{Metric}              \\ \\ 
\shortstack{Left\\Sensor} & \shortstack{Right\\Sensor} & \shortstack{Disparity\\Maps} & & EER & \shortstack{FNR\\FPR=0.05} & \shortstack{FNR\\FPR=0.01} \\
 \cmidrule(r){1-7}
+ &  & + & & 6.82\% & 9.01\% & 24.33\% \\

 & + & + &  &2.23\% & 1.02\% & 4.46\% \\

+ & + &  & & 6.47\% & 7.79\% & 18.89\% \\

+ & + & + & & 1.71\%  & 0.64\% & 2.77\% \\

 \bottomrule
\end{tabular}
}
\\
\scriptsize Lower values indicate better performance.
\end{table}

\subsection{2D Attacks Comparison}
In \autoref{tbl:main_results}, we present our experimental results for the Disparity Model on the test benchmark, which includes both live and 2D attacks. Additionally, we provide results for other anti-spoofing methods across various metrics. The best result in each metric is highlighted in bold. We observe that our method achieves better results than other previous approaches in all metrics.

The Facial Features Pairs approach, which utilizes distances between facial landmarks including their sparse disparity information, performs poorly in our challenging domain. This can be attributed to its lack of texture information and the absence of spatial disparity information, which our method includes. Texture-based approaches that directly use the sensors data show better performance, with the Multilevel Fusion model outperforming the others with an EER of 4.16\%.

However, our multi-modal method, which incorporates disparity maps in addition to raw sensors data as texture, demonstrates superior performance compared to the Multilevel Fusion model. Specifically, our method improves the EER by 2.45\% and the FNR by 3.35\%, 3.27\%, and 7.94\% at spoof acceptance rates of $\frac{5}{100}$, $\frac{3}{100}$, and $\frac{1}{100}$, respectively. 

Furthermore, the ablation study presented in \autoref{tbl:ablation} highlights the significance of disparity maps in the disparity model. The results indicate that integrating disparity maps with the raw data from the right sensor (that includes an IR channel) outperforms the integration with the data from the left sensor. However, the combination of all components results in the highest performance.

Moreover, we analyzed the weights from the first convolutional layer of the Disparity model to assess the impact of each input type on the model’s output. The disparity maps reached a maximum of value of 480, while the raw sensor data values could attain up to 1024. To ensure a fair comparison, the kernel weights of the disparity maps were scaled by a factor of ($\frac{480}{1024}$). The normalized histogram of the first layer’s weights, as shown in \autoref{fig:weights}, indicates that the weight values corresponding to the disparity maps are distributed similarly to those of the raw sensor data. This suggests that both input types contribute equally to the model’s decision making.

\begin{table}[ht!]
\centering
 \caption{Model Ensemble results}
  \label{tbl:ensemble_results}
{%
\begin{tabular}{lcccccc}
\toprule
\shortstack{Spoof Attacks\\Means} & EER &\shortstack{FNR\\FPR=0.05} & \shortstack{FNR\\FPR=0.01} & \shortstack{Test Set\\Size} \\
 \cmidrule(r){1-5}
Paper A4 & 2.10\% & 0.8\% &  6\% & 43132\\
Paper A3 & 1.48\%  & 0.63\% & 1.82\% & 45242\\
TV screen & 1.39\% & 0.32\% & 1.8\% & 41447\\
Desktop screen & 0.86\% & 0.18\% & 0.65\% & 26696\\
Laptop screen & 2.67\% & 1.35\% & 7.58\% & 35531\\
Tablet screen & 0.74\% & 0.11\% & 0.52\% & 37332\\
Mobile Phone screen & 0.75\% & 0.1\% & 0.57\% & 26745\\
Paper on Face & 2.56\% & 1.34\% & 3.65\% & 26015\\
Latex Masks & 3.57\% & 3.05\% & 6.11\% & 26119\\
 \cmidrule(r){1-5}
All & 2.04\% & 0.79\% & 3.83\% & 113403\\

 \bottomrule
\end{tabular}
}
\\
\scriptsize Lower values indicate better performance.
\end{table}

\subsection{Model Ensemble}
Our ensemble comprises three models: the Left Model, the Right Model, and the Disparity Model. This ensemble was evaluated on the complete test set, which includes both 2D and 3D attacks. As presented in \autoref{tbl:ensemble_results}, the overall benchmark, encompassing all types of spoof attacks, yielded an Equal Error Rate (EER) of 2.04\%, while for a false acceptance rate of $\frac{1}{100}$, FNR is 3.83\%. Among the various spoof attack means, the tablet screen, mobile phone screen, and desktop screen exhibited the lowest EER of 0.74\%, 0.75\%, and 0.86\%, respectively. In contrast, the challenging 3D Latex Masks, had the highest EER of 3.57\%.

\section{Conclusion}
A crucial component in anti-spoofing systems is depth information, which is essential for addressing 2D attacks that exhibit unique 3D structures. However, systems that provide depth information are typically expensive and challenging to scale across large, distributed edge device networks. Conversely, non-calibrated systems are more cost-effective but lack depth information.

In this work, we introduce a novel method that overcomes this trade-off by leveraging facial feature knowledge to create proxy-depth maps from pairs of sensors. Our pipeline involves acquiring and processing images from both sensors, detecting faces, and extracting facial landmarks from each image. Using these pairs of facial landmarks, we calculate disparity in each dimension and interpolate all disparity points to produce a disparity map containing relative depth information.

We demonstrate the effectiveness of our method by incorporating the disparity maps as an additional input to a FAS system for edge devices. Our multi-modality anti-spoofing model provides a robust solution for FAS systems in real-life scenarios, outperforming other methods. Specifically, our method was tested on a large benchmark with approximately 100,000 samples, achieving an Equal Error Rate (EER) of 1.75\%, which is 2.45\% lower than the best compared method results. The Importance of the Disparity maps, is reflected in the conducted ablation study and in the CNN weights analysis as well. Furthermore, we show the effectiveness of a three-model ensemble over a larger benchmark that includes 3D attacks, achieving an EER of 2.04\% and an FNR of 3.83\% at a $\frac{1}{100}$ miss acceptance rate.

Overall, our work provides state-of-the-art (SOTA) results for non-calibrated sensor systems in the anti-spoofing task. The study was performed with data collected from the Intel® RealSense™ ID Solution F455 device, but our methodology can be performed on other types of non-calibrated sensors as well.

\section{Acknowledgments}
The authors would like to thank Nelkenbaum Ilya for developing sensor-specific Face-Detection module. At the same time the authors would like to thank Rubinstein Ron for developing the sensor-specific facial landmarks extraction model.   

\bibliographystyle{unsrtnat}
\bibliography{references}

\begin{thebibliography}{29}
\providecommand{\natexlab}[1]{#1}
\providecommand{\url}[1]{\texttt{#1}}
\expandafter\ifx\csname urlstyle\endcsname\relax
  \providecommand{\doi}[1]{doi: #1}\else
  \providecommand{\doi}{doi: \begingroup \urlstyle{rm}\Url}\fi

\bibitem[Zhang et~al.(2022{\natexlab{a}})Zhang, Sun, Wu, and Luo]{CasiaSurf}
Licheng Zhang, Nan Sun, Xihong Wu, and Dingsheng Luo.
\newblock Advanced face anti-spoofing with depth segmentation.
\newblock In \emph{2022 International Joint Conference on Neural Networks (IJCNN)}, pages 1--6, 2022{\natexlab{a}}.
\newblock \doi{10.1109/IJCNN55064.2022.9892826}.

\bibitem[Steiner et~al.(2016)Steiner, Kolb, and Jung]{steiner2016reliable}
Holger Steiner, Andreas Kolb, and Norbert Jung.
\newblock Reliable face anti-spoofing using multispectral swir imaging.
\newblock In \emph{2016 international conference on biometrics (ICB)}, pages 1--8. IEEE, 2016.

\bibitem[Zhang et~al.(2019)Zhang, Wang, Liu, Zhao, Wan, Escalera, Shi, Wang, and Li]{zhang2019dataset}
Shifeng Zhang, Xiaobo Wang, Ajian Liu, Chenxu Zhao, Jun Wan, Sergio Escalera, Hailin Shi, Zezheng Wang, and Stan~Z Li.
\newblock A dataset and benchmark for large-scale multi-modal face anti-spoofing.
\newblock In \emph{Proceedings of the IEEE/CVF Conference on Computer Vision and Pattern Recognition}, pages 919--928, 2019.

\bibitem[Jiang et~al.(2019)Jiang, Liu, and Zhou]{jiang2019multilevel}
Fangling Jiang, Pengcheng Liu, and Xiangdong Zhou.
\newblock Multilevel fusing paired visible light and near-infrared spectral images for face anti-spoofing.
\newblock \emph{Pattern recognition letters}, 128:\penalty0 30--37, 2019.

\bibitem[Lin et~al.(2024)Lin, Wang, Cai, Liu, Fu, Tang, Yu, and Kot]{lin2024suppress}
Xun Lin, Shuai Wang, Rizhao Cai, Yizhong Liu, Ying Fu, Wenzhong Tang, Zitong Yu, and Alex Kot.
\newblock Suppress and rebalance: Towards generalized multi-modal face anti-spoofing.
\newblock In \emph{Proceedings of the IEEE/CVF Conference on Computer Vision and Pattern Recognition}, pages 211--221, 2024.

\bibitem[Hu et~al.(2018)Hu, Shen, and Sun]{hu2018squeeze}
Jie Hu, Li~Shen, and Gang Sun.
\newblock Squeeze-and-excitation networks.
\newblock In \emph{Proceedings of the IEEE conference on computer vision and pattern recognition}, pages 7132--7141, 2018.

\bibitem[Kendall et~al.(2015)Kendall, Badrinarayanan, and Cipolla]{kendall2015bayesian}
Alex Kendall, Vijay Badrinarayanan, and Roberto Cipolla.
\newblock Bayesian segnet: Model uncertainty in deep convolutional encoder-decoder architectures for scene understanding.
\newblock \emph{arXiv preprint arXiv:1511.02680}, 2015.

\bibitem[Yu et~al.(2020{\natexlab{a}})Yu, Qin, Li, Wang, Zhao, Lei, and Zhao]{yu2020multi}
Zitong Yu, Yunxiao Qin, Xiaobai Li, Zezheng Wang, Chenxu Zhao, Zhen Lei, and Guoying Zhao.
\newblock Multi-modal face anti-spoofing based on central difference networks.
\newblock In \emph{Proceedings of the IEEE/CVF Conference on Computer Vision and Pattern Recognition Workshops}, pages 650--651, 2020{\natexlab{a}}.

\bibitem[Yu et~al.(2020{\natexlab{b}})Yu, Zhao, Wang, Qin, Su, Li, Zhou, and Zhao]{yu2020searching}
Zitong Yu, Chenxu Zhao, Zezheng Wang, Yunxiao Qin, Zhuo Su, Xiaobai Li, Feng Zhou, and Guoying Zhao.
\newblock Searching central difference convolutional networks for face anti-spoofing.
\newblock In \emph{Proceedings of the IEEE/CVF conference on computer vision and pattern recognition}, pages 5295--5305, 2020{\natexlab{b}}.

\bibitem[Atoum et~al.(2017)Atoum, Liu, Jourabloo, and Liu]{atoum2017face}
Yousef Atoum, Yaojie Liu, Amin Jourabloo, and Xiaoming Liu.
\newblock Face anti-spoofing using patch and depth-based cnns.
\newblock In \emph{2017 IEEE international joint conference on biometrics (IJCB)}, pages 319--328. IEEE, 2017.

\bibitem[Wang et~al.(2017)Wang, Nian, Li, Meng, and Wang]{wang2017robust}
Yan Wang, Fudong Nian, Teng Li, Zhijun Meng, and Kongqiao Wang.
\newblock Robust face anti-spoofing with depth information.
\newblock \emph{Journal of Visual Communication and Image Representation}, 49:\penalty0 332--337, 2017.

\bibitem[Zhang et~al.(2022{\natexlab{b}})Zhang, Sun, Wu, and Luo]{zhang2022advanced}
Licheng Zhang, Nan Sun, Xihong Wu, and Dingsheng Luo.
\newblock Advanced face anti-spoofing with depth segmentation.
\newblock In \emph{2022 International Joint Conference on Neural Networks (IJCNN)}, pages 1--6. IEEE, 2022{\natexlab{b}}.

\bibitem[Zhao et~al.(2017)Zhao, Shi, Qi, Wang, and Jia]{zhao2017pyramid}
Hengshuang Zhao, Jianping Shi, Xiaojuan Qi, Xiaogang Wang, and Jiaya Jia.
\newblock Pyramid scene parsing network.
\newblock In \emph{Proceedings of the IEEE conference on computer vision and pattern recognition}, pages 2881--2890, 2017.

\bibitem[Yang et~al.(2019)Yang, Luo, Bao, Gao, Gong, Zheng, Li, and Liu]{yang2019face}
Xiao Yang, Wenhan Luo, Linchao Bao, Yuan Gao, Dihong Gong, Shibao Zheng, Zhifeng Li, and Wei Liu.
\newblock Face anti-spoofing: Model matters, so does data.
\newblock In \emph{Proceedings of the IEEE/CVF conference on computer vision and pattern recognition}, pages 3507--3516, 2019.

\bibitem[Li et~al.(2016)Li, Komulainen, Zhao, Yuen, and Pietik{\"a}inen]{li2016generalized}
Xiaobai Li, Jukka Komulainen, Guoying Zhao, Pong-Chi Yuen, and Matti Pietik{\"a}inen.
\newblock Generalized face anti-spoofing by detecting pulse from face videos.
\newblock In \emph{2016 23rd International Conference on Pattern Recognition (ICPR)}, pages 4244--4249. IEEE, 2016.

\bibitem[Ge et~al.(2020)Ge, Tu, Ai, Luo, Ma, and Xie]{ge2020face}
Hao Ge, Xiaoguang Tu, Wenjie Ai, Yao Luo, Zheng Ma, and Mei Xie.
\newblock Face anti-spoofing by the enhancement of temporal motion.
\newblock In \emph{2020 2nd International Conference on Advances in Computer Technology, Information Science and Communications (CTISC)}, pages 106--111. IEEE, 2020.

\bibitem[Greff et~al.(2016)Greff, Srivastava, Koutn{\'\i}k, Steunebrink, and Schmidhuber]{greff2016lstm}
Klaus Greff, Rupesh~K Srivastava, Jan Koutn{\'\i}k, Bas~R Steunebrink, and J{\"u}rgen Schmidhuber.
\newblock Lstm: A search space odyssey.
\newblock \emph{IEEE transactions on neural networks and learning systems}, 28\penalty0 (10):\penalty0 2222--2232, 2016.

\bibitem[Wu et~al.(2012)Wu, Rubinstein, Shih, Guttag, Durand, and Freeman]{wu2012eulerian}
Hao-Yu Wu, Michael Rubinstein, Eugene Shih, John Guttag, Fr{\'e}do Durand, and William Freeman.
\newblock Eulerian video magnification for revealing subtle changes in the world.
\newblock \emph{ACM transactions on graphics (TOG)}, 31\penalty0 (4):\penalty0 1--8, 2012.

\bibitem[Anthony and Ay(2022)]{anthony2022active}
Peter Anthony and Bet{\"u}l Ay.
\newblock Active face spoof detection using image distortion analysis.
\newblock \emph{Turkish Journal of Science and Technology}, 17\penalty0 (2):\penalty0 435--450, 2022.

\bibitem[Li et~al.(2008)Li, Gunturk, and Zhang]{li2008image}
Xin Li, Bahadir Gunturk, and Lei Zhang.
\newblock Image demosaicing: A systematic survey.
\newblock In \emph{Visual Communications and Image Processing 2008}, volume 6822, pages 489--503. SPIE, 2008.

\bibitem[Virtanen et~al.(2020)Virtanen, Gommers, Oliphant, Haberland, Reddy, Cournapeau, Burovski, Peterson, Weckesser, Bright, {van der Walt}, Brett, Wilson, Millman, Mayorov, Nelson, Jones, Kern, Larson, Carey, Polat, Feng, Moore, {VanderPlas}, Laxalde, Perktold, Cimrman, Henriksen, Quintero, Harris, Archibald, Ribeiro, Pedregosa, {van Mulbregt}, and {SciPy 1.0 Contributors}]{2020SciPy-NMeth}
Pauli Virtanen, Ralf Gommers, Travis~E. Oliphant, Matt Haberland, Tyler Reddy, David Cournapeau, Evgeni Burovski, Pearu Peterson, Warren Weckesser, Jonathan Bright, St{\'e}fan~J. {van der Walt}, Matthew Brett, Joshua Wilson, K.~Jarrod Millman, Nikolay Mayorov, Andrew R.~J. Nelson, Eric Jones, Robert Kern, Eric Larson, C~J Carey, {\.I}lhan Polat, Yu~Feng, Eric~W. Moore, Jake {VanderPlas}, Denis Laxalde, Josef Perktold, Robert Cimrman, Ian Henriksen, E.~A. Quintero, Charles~R. Harris, Anne~M. Archibald, Ant{\^o}nio~H. Ribeiro, Fabian Pedregosa, Paul {van Mulbregt}, and {SciPy 1.0 Contributors}.
\newblock {{SciPy} 1.0: Fundamental Algorithms for Scientific Computing in Python}.
\newblock \emph{Nature Methods}, 17:\penalty0 261--272, 2020.
\newblock \doi{10.1038/s41592-019-0686-2}.

\bibitem[Sandler et~al.(2018)Sandler, Howard, Zhu, Zhmoginov, and Chen]{sandler2018mobilenetv2}
Mark Sandler, Andrew Howard, Menglong Zhu, Andrey Zhmoginov, and Liang-Chieh Chen.
\newblock Mobilenetv2: Inverted residuals and linear bottlenecks.
\newblock In \emph{Proceedings of the IEEE conference on computer vision and pattern recognition}, pages 4510--4520, 2018.

\bibitem[Sensoy et~al.(2018)Sensoy, Kaplan, and Kandemir]{sensoy2018evidential}
Murat Sensoy, Lance Kaplan, and Melih Kandemir.
\newblock Evidential deep learning to quantify classification uncertainty.
\newblock \emph{Advances in neural information processing systems}, 31, 2018.

\bibitem[Achrack et~al.(2020)Achrack, Kellerman, and Barzilay]{achrack2020multi}
Omer Achrack, Raizy Kellerman, and Ouriel Barzilay.
\newblock Multi-loss sub-ensembles for accurate classification with uncertainty estimation.
\newblock \emph{arXiv preprint arXiv:2010.01917}, 2020.

\bibitem[Neumann et~al.(2018)Neumann, Zisserman, and Vedaldi]{neumann2018relaxed}
Lukas Neumann, Andrew Zisserman, and Andrea Vedaldi.
\newblock Relaxed softmax: Efficient confidence auto-calibration for safe pedestrian detection.
\newblock 2018.

\bibitem[Abadi(2016)]{abadi2016tensorflow}
Mart{\'\i}n Abadi.
\newblock Tensorflow: learning functions at scale.
\newblock In \emph{Proceedings of the 21st ACM SIGPLAN international conference on functional programming}, pages 1--1, 2016.

\bibitem[Kingma(2014)]{kingma2014adam}
Diederik~P Kingma.
\newblock Adam: A method for stochastic optimization.
\newblock \emph{arXiv preprint arXiv:1412.6980}, 2014.

\bibitem[Redmon(2016)]{redmon2016you}
J~Redmon.
\newblock You only look once: Unified, real-time object detection.
\newblock In \emph{Proceedings of the IEEE conference on computer vision and pattern recognition}, 2016.

\bibitem[Ronneberger et~al.(2015)Ronneberger, Fischer, and Brox]{ronneberger2015u}
Olaf Ronneberger, Philipp Fischer, and Thomas Brox.
\newblock U-net: Convolutional networks for biomedical image segmentation.
\newblock In \emph{Medical image computing and computer-assisted intervention--MICCAI 2015: 18th international conference, Munich, Germany, October 5-9, 2015, proceedings, part III 18}, pages 234--241. Springer, 2015.

\end{thebibliography}


\clearpage
\appendix
\section*{Appendix}
\addcontentsline{toc}{section}{Appendix}

\begin{figure*}
    \centering
    \includegraphics[width=0.95\textwidth]{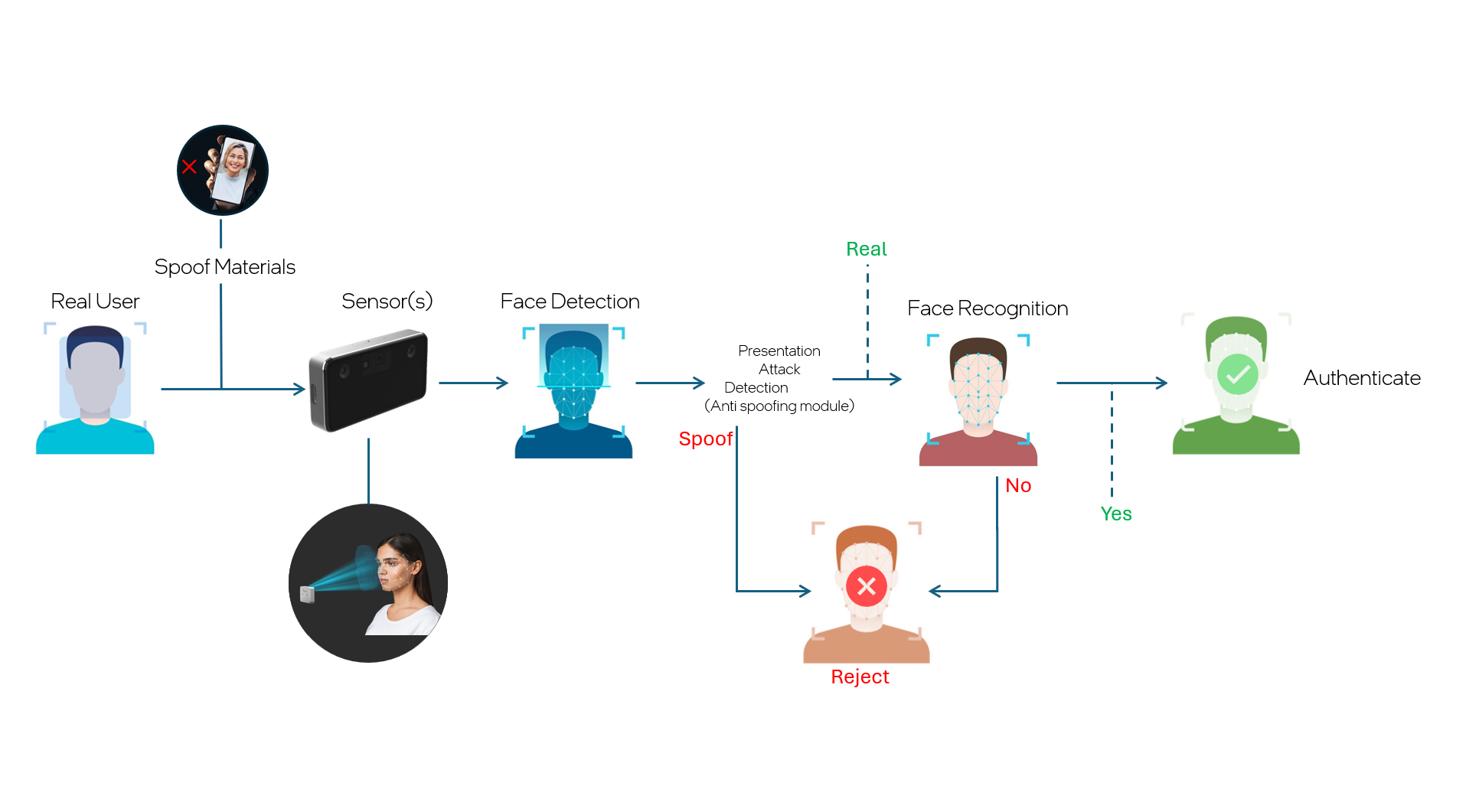}
    \captionsetup{width=0.8\textwidth}
    \caption{Schematic view of typical Face Recognition pipeline}
    \label{fig:fa_pipe}
\end{figure*}

\section{Face Authentication pipeline}
\label{app:fa_pipeline}
A typical pipeline for Face Authentication (FA) begins with image acquisition. The face within the image is detected using a dedicated face detector model, which predicts the bounding box around the face. Once the face is detected, by utilizing a dedicated anti-spoof model, it undergoes discrimination based on various features to determine whether it is a live face or a spoof attack.
If the anti-spoofing step identifies the processed image as a spoof, the pipeline is terminated. Otherwise, representative features are extracted from the image in the final step. These image features are then compared to other features representing a specific identity to ascertain whether the captured image corresponds to that specific identity. \autoref{fig:fa_pipe} illustrates the FA pipeline.

\begin{figure*}
    \begin{center}
    \begin{subfigure}[b]{0.45\textwidth}  \includegraphics[width=0.9\linewidth]{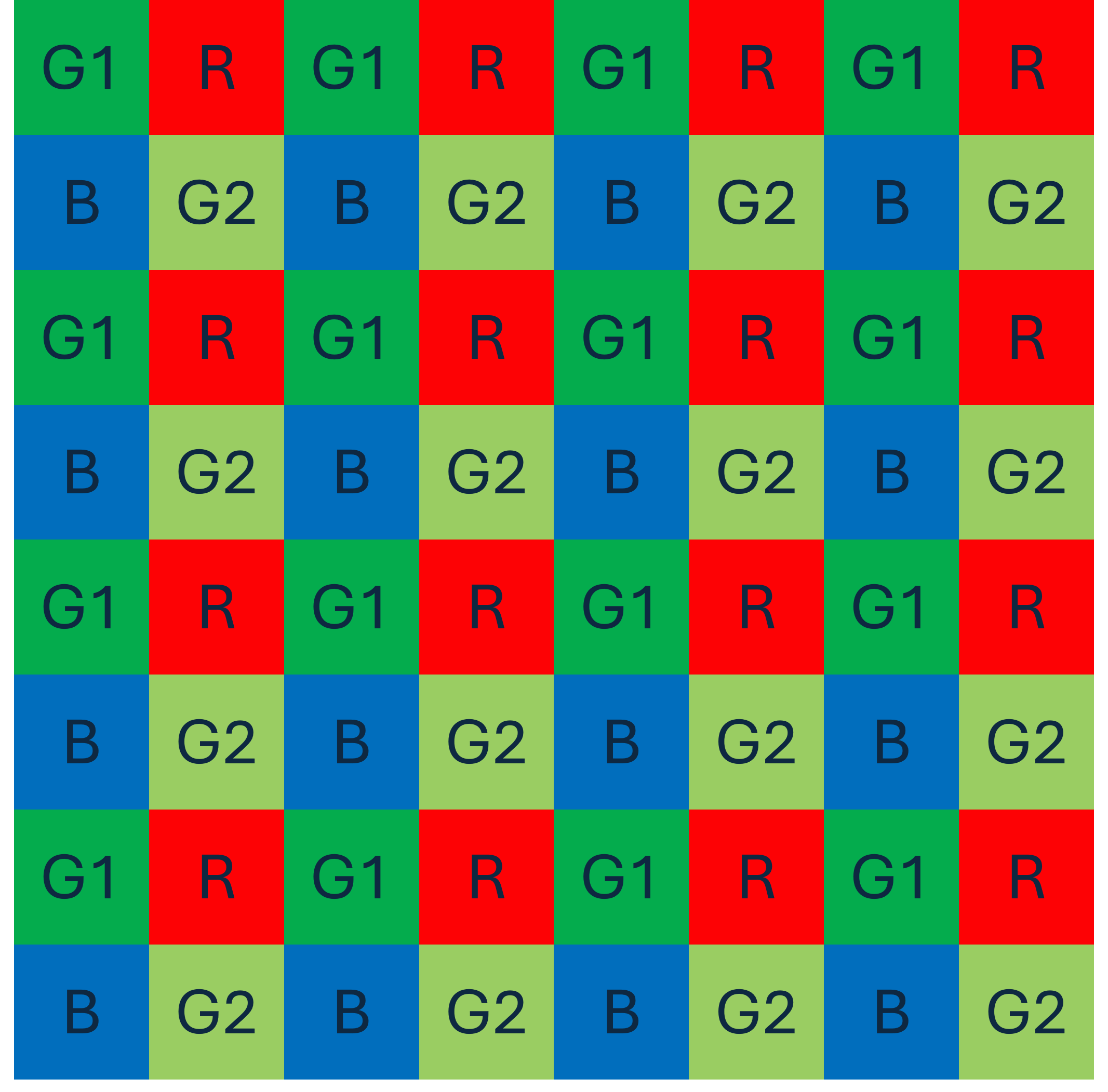}
        \caption{Raw Bayer input}
    \end{subfigure}
    \begin{subfigure}[b]{0.45\textwidth}
        \includegraphics[width=0.9\linewidth]{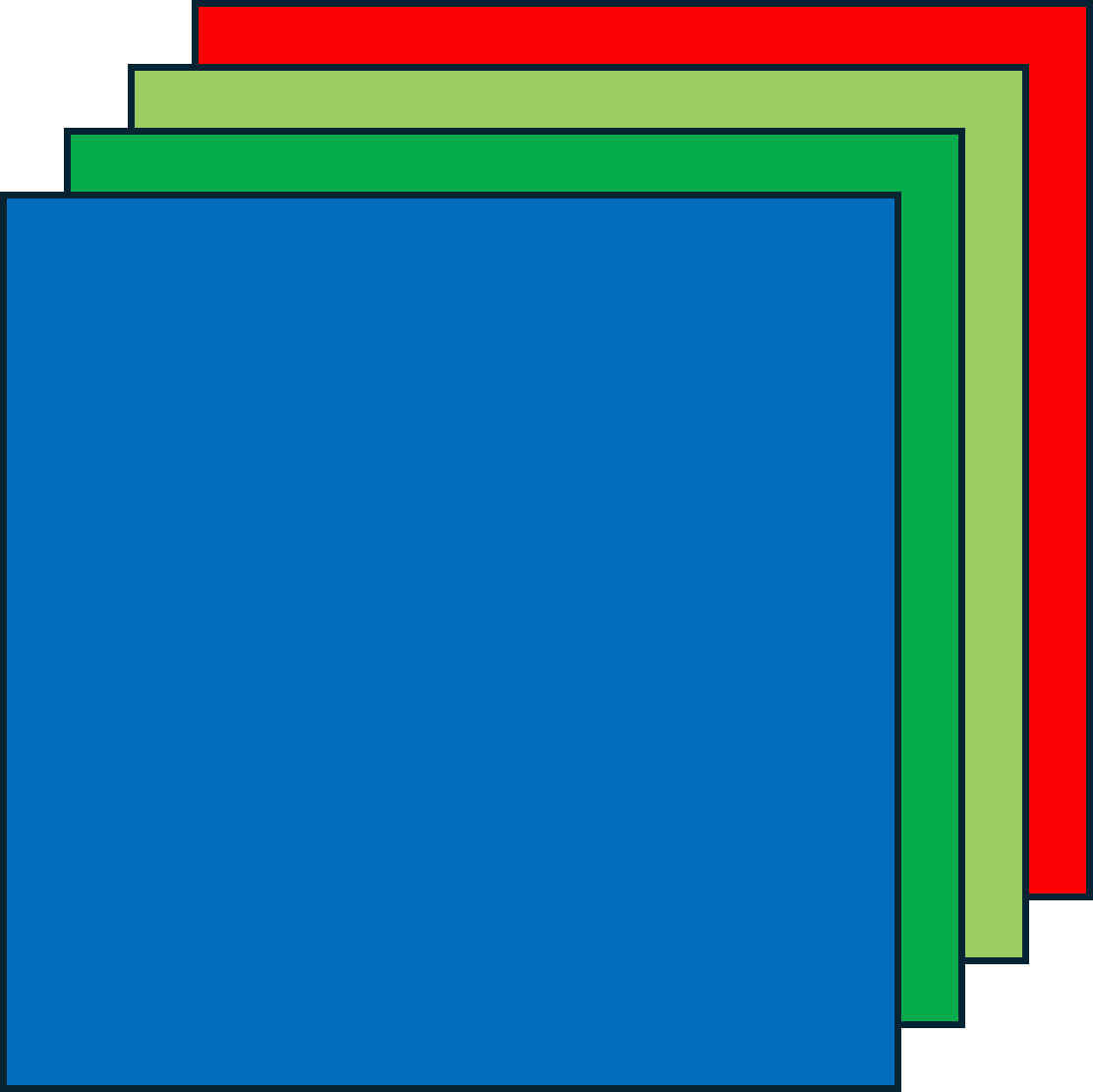}
        \caption{Conversion into 4 channels (BGGR)}
    \end{subfigure}
        \caption{Conversion process of Bayer pattern into four Channels BGGR image. The spatial resolution is reduced by a factor of two in each dimension to create four channels image instead of one.}
        \label{fig:bayer_vs_bggr}
    \end{center}
\end{figure*}

\section{Raw Bayer Pattern Process}
\label{app:bayer_vs_bggr}
In our pipeline the raw Bayer pattern of the sensor is processed directly (without demosaicing). The Bayer pattern of each sensor consists of 2x2 pixels, where each pixel within this pattern is considered as a single channel, resulting in a four-channel input for each sensor. An example for left sensor pattern is presented in \autoref{fig:bayer_vs_bggr}.

\section{Supplementary Models}
\label{app:Supplementary_Models}

As part of the pipeline, each sensor’s data is cropped around the face using a dedicated face detector (FD) model. This FD model processes a single channel from the sensor (green) and predicts the coordinates of a bounding box around the face, along with a confidence score. The architecture of the FD model is based on YOLO, and it refines its initial anchors during the detection process (\cite{redmon2016you}). The model is trained on grayscale images from approximately 200,000 subjects and is subsequently fine-tuned with around 5,000 annotated samples collected from the Intel® RealSense™ ID Solution F455.
Another integral component of our pipeline is the Facial Landmarks Extractor (LE). Similar to the Face Detector (FD), the LE processes a single channel and predicts the coordinates of 45 facial landmarks: 5 for each eyebrow, 7 for each eye, 9 for the nose, and 12 for the lips. The architecture of the LE is a variation of U-Net (\cite{ronneberger2015u}), followed by a CNN regressor to predict the 45 2D coordinates. It is trained using approximately 240,000 images rendered from 3D scans under various environmental conditions, head poses, and lighting scenarios. Additionally, around 160,000 samples containing RGB images and images captured with the Intel® RealSense™ ID Solution F455 were labeled and incorporated into the training process.

\end{document}